\documentclass{article}
\usepackage{spconf,amsmath,graphicx}
\usepackage{multicol}
\usepackage{multirow}
\usepackage{color}
\usepackage{cleveref}

\usepackage{enumitem}
\usepackage{etoolbox}
\usepackage{makecell}
\usepackage{wrapfig}
\usepackage{setspace}
\usepackage{floatflt}
\usepackage{float}
\usepackage{caption}
\usepackage{subcaption}

\usepackage{tikz}
\usepackage{pgfplots}
\usetikzlibrary{positioning,chains,shapes.geometric, calc, shadows, shapes.misc}

\pgfplotsset{compat=1.16}

\title{Robust Knowledge Distillation from RNN-T Models With Noisy Training Labels Using Full-Sum Loss}

\name{Mohammad Zeineldeen$^{1*}$\thanks{$^*$Work performed during an internship at Google.}, Kartik Audhkhasi$^2$, Murali Karthick Baskar$^2$, Bhuvana Ramabhadran$^2$}
\address{
$^1$Human Language Technology and Pattern Recognition, Computer Science Department, \\
RWTH Aachen University, 52074 Aachen, Germany \\
$^2$Google LLC, New York
} 
\renewrobustcmd{\bfseries}{\fontseries{b}\selectfont}
\renewrobustcmd{\boldmath}{}
\newrobustcmd{\B}{\bfseries}

\newsavebox\CBox

\makeatletter

\renewcommand{\section}{\@startsection
   {section}%
   {1}%
   {}%
   {-0.4\baselineskip}%
   {0.2\baselineskip}%
   {}}%

\renewcommand{\subsection}{\@startsection
  {subsection}%
  {2}%
  {}%
  {-0.1\baselineskip}%
  {0.1\baselineskip}%
  {}}%

\renewcommand{\subsubsection}{\@startsection
  {subsubsection}%
  {3}%
  {}%
  {-0.2\baselineskip}%
  {0.2\baselineskip}%
  {}}%

\g@addto@macro\normalsize{%
  \setlength\abovedisplayskip{3pt plus 2pt minus 1pt}
  \setlength\belowdisplayskip{3pt plus 2pt minus 1pt}
  \setlength\abovedisplayshortskip{2pt plus 2pt minus 1pt}
  \setlength\belowdisplayshortskip{2pt plus 2pt minus 1pt}
}

\begin{document}
%
\maketitle
\begin{abstract}
This work studies knowledge distillation (KD) and addresses its constraints for recurrent neural network transducer (RNN-T) models.
In hard distillation, a teacher model transcribes large amounts of unlabelled speech to train a student model.
Soft distillation is another popular KD method that distills the output logits of the 
teacher model.
Due to the nature of RNN-T alignments, applying soft distillation between
RNN-T architectures having different posterior distributions is challenging.
In addition, bad teachers having high word-error-rate (WER) reduce the efficacy of KD.
We investigate how to effectively distill knowledge from variable
quality ASR teachers, which has not been studied before to the best of our knowledge.
We show that a sequence-level KD, full-sum distillation, outperforms other distillation methods for RNN-T models, especially for bad teachers. We also propose a variant of full-sum distillation that distills the
sequence discriminative knowledge of the teacher leading to further improvement in
WER.
We conduct experiments on public datasets namely SpeechStew and LibriSpeech, 
and on in-house production data.
\end{abstract}
\section{Introduction \& Related Work}
\label{sec:intro}


%
Training high-performance end-to-end automatic speech recognition (ASR) systems 
such as recurrent neural network transducer (RNN-T) \cite{rnnt-paper} heavily depends on 
the amount and the quality of the transcribed training data.
It is usually difficult and very expensive to collect high-quality human
transcription.

%
Knowledge distillation (KD) \cite{kd-paper} is a method to transfer knowledge 
from a teacher model to a (smaller) student model.
The teacher model generates pseudo labels using unsupervised
data for training a student model.
%
%
However, the quality of the pseudo labels depends on the quality of the 
teacher model where a bad teacher, i.e with a high word-error-rate (WER), can generate 
noisy pseudo labels which do not help in training good student models.
%
%
To the best of our knowledge, there has been no prior work that investigates the impact of 
bad teachers for KD in the context of ASR.
\cite{kd-bad-good} shows theoretically and empirically that distilling from a pool of bad teachers (randomly selected) helps to learn a better student model.
\cite{SauB16,LukasikBMK20} proposes a simpler idea which is to add noise to the teacher’s 
logits to simulate the idea of training with multiple bad teachers.
Both \cite{kd-bad-good,SauB16} can be seen as regularization methods.
There has also been related work in the context of adversarial label learning \cite{ArachieH19,ArachieH21} where students are trained to minimize the 
error caused by noisy labels generated by bad teachers. 
However, all these works use simple binary classifiers and focus more on 
theoretical analysis.

In addition, it is common to apply KD at the level of output logits \cite{kd-paper}.
Soft distillation can be used to distill the RNN-T alignments of the teacher model.
However, this method is challenging when the teacher and student models
have different alignments such as distilling knowledge from a non-causal 
teacher to a causal student.
To fix this issue, \cite{dongseong-slt, YangLW22b} shift the teacher alignments to the right when applying soft distillation since the causal student would emit labels later because of missing future context.
However, it is a heuristic solution that requires finding the frame shift and increases the causal latency.
In addition, when using soft distillation, the teacher and student 
models must have the same time dimension which limits the ability
to train student models with higher time reduction for 
reducing recognition latency. \cite{TakashimaLK18} investigates different KD methods for connectionist temporal classification models \cite{GravesFGS06} including sequence-level KD, which has not been studied for RNN-T models.

In this paper, we investigate how to effectively distill knowledge from varying
quality RNN-T teachers including bad teachers which have not been studied before.
We apply full-sum distillation, which is a sequence-level KD method that distills
the sequence posterior probabilities, for the first
time for RNN-T models using various loss functions.
We also propose a variant of full-sum distillation which distills the
sequence discriminative knowledge of the teacher model that leads to further
improvements.
We show that full-sum distillation is robust towards discrepancies of RNN-T alignments
between the teacher and student models and that it scales when applied on
our in-house data.

\section{RNN-T Model}
\label{sec:rnnt}

In this work, we focus on the standard RNN-T model \cite{rnnt-paper}.
Let $X$ denote the acoustic feature sequence of a speech utterance of length $T$.
Let $Y$ denote the output label sequence (e.g characters) of length $U$.
Then, the sequence posterior probability is defined as:
\begin{equation}
\label{eq:rnnt_seq_prob}
    P(Y \vert X) = \sum_{a \in \beta^{-1}(Y)} P(a \vert X)   
\end{equation}
where $a$ belongs to the set of all possible alignments of $Y$ consisting of output labels and a special blank label.
$\beta$ is a mapping function that maps an alignment sequence $a$ to an output label sequence $Y$ by removing blank labels.
The RNN-T training loss is given by the negative log sequence posterior probability:
$
    \mathcal{L}_{\text{RNN-T}} = - \mathrm{log} P(Y \vert X).
$
This is also known as the full-sum (FS) loss.
The probability $P(Y|X)$ is computed over a lattice of dimension $T \times U$.
At each position $(t,u)$ in the lattice, features from the encoder and the prediction networks of the RNN-T model are fed to a joint network that computes a probability distribution $P(k|t,u)$ for each
output label $k$ including the blank label.
The RNN-T loss can be computed efficiently using forward-backward algorithm \cite{rnnt-paper}.

\section{RNN-T Distillation Methods}
\label{sec:rnnt-distill}

The most common distillation methods are hard distillation and soft distillation \cite{kd-paper}.
Hard distillation uses the pseudo labels generated by a teacher model to 
train a student model.
It is also possible to use a mix of pseudo labels 
and ground truth labels (supervised data).
For RNN-T models, this means that the student model would learn 
the alignment by itself (i.e without any constraints) 
by minimizing the RNN-T loss. 
Soft distillation is applied by matching the posterior probability distributions for each output label $k$ (including blank) of both teacher and student models over the lattice for each position $(t, u)$ \cite{PanchapagesanPC21}.
This can be achieved using Kullback-Leibler (KL) divergence loss as follows:
\begin{equation}
    \mathcal{L}_{\text{Soft-Distill}} = \sum_{(t, u)} \sum_{k} \widetilde{P}(k \vert t, u) \log\Bigg{[}\frac{\widetilde{P}(k \vert t, u)}{P(k \vert t, u)}\Bigg{]}
\end{equation}
where $\widetilde{P}$ and $P$ correspond to the teacher and student probability distributions respectively.
\cite{PanchapagesanPC21} proposes an efficient method to apply soft distillation for RNN-T models by distilling only three posterior probabilities which are for
target output label, blank label, and the rest labels.
This reduces the memory complexity from $O(T \times U \times K)$ to $O(T \times U)$
where $K$ is the vocabulary size.
We use this efficient method for experiments with soft distillation.

\section{Full-sum Distillation Method}
\label{sec:full-sum-distill}

The main motivation behind this work is to utilize a simple yet effective 
KD method that is robust in case of noisy labels and 
when the architecture or design of the student and teacher models differs.
Therefore, we use full-sum (FS) distillation, as a sequence-level KD method, 
that simply distills the FS probabilities between the teacher and student model.
The loss can be defined as:
\begin{equation}
\label{eq:fs_distill}
    \mathcal{L}_{\text{FS-Distill}} = \mathcal{F}(\widetilde{P}(Y|X), P(Y|X))
\end{equation}
where $\mathcal{F}$ denotes the loss function used to minimize the difference between both distributions.
In this work, $\mathcal{F}$ is defined as $\mathrm{L_1}$ loss or 
mean squared error ($\mathrm{MSE}$) loss so that it is symmetric and 
is not impacted by any transformation of its two arguments.
Therefore, we found that in practice using log-space probabilities makes training more stable.
In addition, we can formulate the FS-Distill loss in terms of RNN-T loss as follows:
\begin{equation}
    \mathcal{L}_{\text{FS-Distill}} = \mathcal{F}(-\mathrm{log} \widetilde{P}(Y|X), -\mathrm{log} P(Y|X)) 
\end{equation}
Thus, we only need to compute the RNN-T losses of both teacher and student models to compute FS-Distill loss.
Moreover, the FS-Distill loss can be formulated to distill the 
sequence discriminative knowledge of the teacher model to the student model.
This can be done by distilling the approximated normalized sequence posterior probabilities using an N-best hypotheses list generated
by the teacher model.
This loss is called FS-Norm-Distill loss 
and can be written as follows:
\begin{equation}
    \label{eq:fs-norm}
    \scalebox{1.05}{$
    \mathcal{F} \biggl( \mathrm{log} \frac{\widetilde{P}(Y|X)}{\sum_{Y' \in \mathcal{B}_{\text{N-best}}} \widetilde{P}(Y'|X)},
    \mathrm{log} \frac{P(Y|X)}{\sum_{Y' \in \mathcal{B}_{\text{N-best}}} P(Y'|X)} \biggl)$
    }
\end{equation}
where $Y'$ belongs to the N-best hypotheses list denoted by $\mathcal{B}_\text{N-best}$.

Note also that FS distillation method does not depend on time dimension as compared to soft distillation which means that the teacher and student models 
can have different time subsampling rates.

\section{Experiments}
\label{sec:experiments}

In this section, we present results on different public corpora, 
namely, SpeechStew \cite{speechstew} (\Cref{subsec:speechstew}) 
and LibriSpeech \cite{libri-corpus} (\Cref{subsec:ls}).
For distillation experiments, no dropout or data augmentation is applied to the speech input of the 
teacher model since it was observed that this leads to better performance \cite{dongseong-slt}.
In addition to that, training batches are constructed by sampling $10\%$ 
from the supervised data and $90\%$ from the unsupervised data.
We use a beam size of 8 to generate hypotheses for unsupervised data and then
select the top-1 hypothesis as the target label sequence.
All student models are trained from scratch.
All models use 80-dimensional Log-Mel filterbank features as input
and 1024 wordpieces as output labels.
No language model is used for recognition.

\subsection{SpeechStew Setup}
\label{subsec:speechstew}

We train varying quality teacher models using SpeechStew dataset \cite{speechstew} to better
understand the effect of each distillation method depending on the teacher's quality.
SpeechStew consists of 5K hours and it is a mix of common speech public corpora.
We split the dataset into 2 parts: supervised data consisting of 250 hours ($5\%$) and unsupervised
data consisting of 4.75K hours ($95\%$).
All data is used for all experiments.
We use ConformerL, ConformerM, and ConformerS RNN-T architectures \cite{conformer-paper} 
to train teacher models having a different number of parameters.
In addition, we train on subsets of the supervised training data to increase the WER variance
between different teacher models.
We train 5 different teacher models named as $\mathrm{L}$, $\mathrm{S}$, $\mathrm{M5}$, $\mathrm{L5}$, and $\mathrm{S3}$
where the letter corresponds to which conformer architecture is used and the number represents the percentage of
supervised training data (e.g $\mathrm{M5}$ is ConformerM trained on all supervised data).
Teachers $\mathrm{L}$ and $\mathrm{S}$ are trained using all SpeechStew data.
The student model follows ConformerS architecture and it 
is trained using the supervised data.

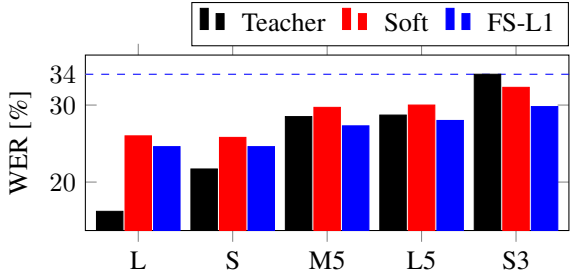
\begin{figure}[t]
\begin{tikzpicture}
\begin{axis}[
    width=0.45\textwidth,
    height=0.22\textwidth,
	ylabel={WER [\%]},
    ybar=2*\pgflinewidth,
    enlargelimits=0.14,
    legend cell align=left,
    legend style={
        at={(1.0,1.05)},
        anchor=south east,
        column sep=1ex,
        legend columns=-1,
        nodes={scale=1.0, transform shape}
    },
    extra y ticks=34.0,
    extra y tick style={
        ymajorgrids=true,
        ytick style={
            /pgfplots/major tick length=0pt,
        },
        grid style={
            blue,
            dashed,
            /pgfplots/on layer=axis foreground,
        },
    },
    xtick={1, 2, 3, 4, 5},
    xticklabels={{L}, {S}, {M5}, {L5}, {S3}}
]

\addplot[color=black,fill=black]
    coordinates {
    (1, 16.2)
    (2, 21.7)
    (3, 28.5)
    (4, 28.7)
    (5, 34.0)
    };

\addplot[color=red,fill=red]
    coordinates {
    (1, 26.0)  
    (2, 25.8)  
    (3, 29.7)  
    (4, 30.0)  
    (5, 32.3)  
    };
    
\addplot[color=blue,fill=blue]
    coordinates {
    (1, 24.6)  
    (2, 24.6)  
    (3, 27.3)  
    (4, 28.0)  
    (5, 29.8)  
    };

\legend{Teacher, Soft, FS-L1}

\end{axis}
\end{tikzpicture}
\caption{
\it Comparison between soft and full-sum distillation using varying quality teachers.
The horizontal dashed line is the WER of the student model.
WERs are computed as the average WER on IHM and SDM1 dev sets of AMI dataset. }
\label{fig:speechstew-soft-vs-fs}
\end{figure}

\subsubsection{Distillation Results}

\begin{table}[t]
    \centering
    \begin{tabular}{|c|c|c|c|c|c|}
        \hline
        \multirow{3}{*}{Model} & \multirow{3}{*}{Teacher} & \multicolumn{2}{c|}{IHM} & \multicolumn{2}{c|}{SDM} \\ \cline{3-6}
                               & & dev & eval & dev & eval \\ \hline
                Student & - & 23.3 & 24.5 & 44.7 & 48.6 \\ \hline \hline
                Hard & \multirow{3}{*}{L} & 24.4 & 24.1 & 38.4 & 40.9 \\ \cline{1-1} \cline{3-6}
                Hard + Soft & & 17.9 & 17.9 & 35.3 & 38.4 \\ \cline{1-1} \cline{3-6}
                Soft & & 17.2 & 17.1 & 34.8 & 38.1 \\ \hline
                Hard & \multirow{3}{*}{S3} & 27.6 & 28.2 & 42.9 & 45.9 \\ \cline{1-1} \cline{3-6}
                Hard + Soft & & 22.3 & 23.1 & 42.3 & 46.1 \\ \cline{1-1} \cline{3-6}
                Soft & & 22.1 & 22.9 & 42.4 & 46.3 \\ \hline
    \end{tabular}
    \caption{\it Comparison between soft and hard distillation.}
    \label{tab:speechstew-hard}
\end{table}

As our aim is to improve knowledge distillation when using bad or high-WER teacher models,
we report results on the AMI dataset \cite{AMI-Corpus} 
since it is noisy and considered a difficult task.
\Cref{tab:speechstew-hard} shows the results of applying hard distillation using the
strongest teacher $\mathrm{L}$ and the weakest teacher $\mathrm{S3}$.
We can observe that even with teacher $\mathrm{L}$, hard distillation is much worse than
using soft distillation.
In addition, WER increases when using the weakest teacher $\mathrm{S3}$.
The reason behind this is that the pseudo labels generated by such teacher models are very noisy, especially on AMI which requires utilizing other distillation methods.
\Cref{fig:speechstew-soft-vs-fs} shows a comparison between
soft distillation and FS distillation when using
varying quality teachers.
We use L1 loss for FS distillation.
First, we can observe that the quality of the teacher
model has a significant effect on improving the WER
of the student model.
FS distillation outperforms soft distillation for all teachers.
The student model outperforms the teacher models $\mathrm{M5}$, $\mathrm{L5}$, and $\mathrm{S3}$ when using FS distillation whereas with soft distillation 
it only outperforms $\mathrm{S3}$.
This shows the robustness of FS distillation method.

\subsubsection{Comparison between L1 and MSE loss for Full-sum Distillation}

\begin{table}[t]
    \centering
    \begin{tabular}{|c|c|c|c|c|c|}
    \hline
     \multirow{2}{*}{$\mathcal{F}$} & \multirow{2}{*}{Norm} & \multicolumn{2}{c|}{IHM} & \multicolumn{2}{c|}{SDM} \\ \cline{3-6}
                             & & dev & eval & dev & eval \\ \hline
                    \multirow{2}{*}{MSE} & No & 17.7 & 17.5 & 34.4 & 37.6 \\ \cline{2-6}
                    & Yes & 16.7 & 16.5 & 33.5 & 36.8 \\ \hline
                    \multirow{2}{*}{L1} & No & 16.1 & 16.4 & 33.1 & 36.2 \\ \cline{2-6}
                    & Yes & \textbf{15.8} & \textbf{15.7} & \textbf{32.3} & \textbf{35.5} \\ \hline
    \end{tabular}
    \caption{\it Comparison between using L1 and MSE loss functions for FS and FS-Norm distillation for teacher $\mathrm{L}$.}
    \label{tab:mse_vs_l1}
\end{table}


We investigate using two different losses for FS distillation: $\mathrm{L1}$
loss and $\mathrm{MSE}$ loss.
We select the strongest teacher $\mathrm{L}$ for
distillation experiments and present the results in \Cref{tab:mse_vs_l1}.
We can observe that using $\mathrm{L1}$ loss gives much better performance in terms of
WER compared to $\mathrm{MSE}$ loss.
We argue that the training convergence is affected by outliers when using $\mathrm{MSE}$
loss.
To analyze this, we plot (plot is missing due to limited space) the distillation loss value using 100 segments from AMI dataset.
For the case of $\mathrm{MSE}$, there are many outliers, and the distillation
loss value is quite large while this is not the case when using $\mathrm{L1}$.
In addition, if we apply approximated normalization as described in
\Cref{eq:fs-norm}, then we do not observe outliers anymore which
could explain also why using FS-Norm variant helps (more details in \Cref{subsubsec:fs-norm}).

\subsubsection{Full-sum Norm Distillation}
\label{subsubsec:fs-norm}

Furthermore, we conduct experiments using FS-Norm variant
(\Cref{eq:fs-norm})
and the results are shown in \Cref{tab:mse_vs_l1}.
We use the strongest teacher $\mathrm{L}$ for distillation.
Applying normalization further improves the WER of the student model
especially when using MSE loss since it reduces outliers. 

\subsection{LibriSpeech Setup}
\label{subsec:ls}


We conduct experiments on LibriSpeech ($\mathrm{LS}$) 960 hours \cite{libri-corpus} 
and LibriLight ($\mathrm{LL}$) 60k hours \cite{librilight}.
$\mathrm{LL}$ consists of unlabeled data which is the main target data for distillation.
The teacher model is a non-causal w2v-BERT XL Conformer model following this setup \cite{dongseong-slt}.
It has 600M parameters.
It is pretrained using w2v-BERT \cite{w2v-bert} on $\mathrm{LL}$ dataset
and then iterative training is applied using offline pseudo labels to further improve 
the performance.
The pseudo labels used were generated by a w2v-BERT XXL model \cite{w2v-bert} having 1B parameters.
The WERs [\%] of w2v-BERT XL and w2v-BERT XXL teacher models are $1.3/2.5/1.4/2.6$ and $1.4/2.4/1.4/2.5$ on dev-clean, dev-other, test-clean, and test-other respectively.
The non-causal student model is based on the ConformerL architecture \cite{conformer-paper}.
It has 120M parameters.
The causal student model uses the same architecture but with causal
conformer blocks where 65 frames are used as past context for self-attention modules 
and no future context.
SpecAugment \cite{specaug} is applied for data augmentation using 
the same hyperparameters as \cite{dongseong-slt}.

\subsubsection{Causal/Non-causal Distillation Results}

\begin{figure}[t]
\begin{tikzpicture}
\begin{axis}[
    width=0.45\textwidth,
    height=0.22\textwidth,
    ylabel={WER [\%]},
    xlabel={N},
    xtick={0,1,2,3,4,5,6,7,8,9,10},
    ytick={10, 15},
    extra y ticks=5.7,
    extra y tick style={
        ymajorgrids=true,
        ytick style={
            /pgfplots/major tick length=0pt,
        },
        grid style={
            blue,
            dashed,
            /pgfplots/on layer=axis foreground,
        },
    },
]

\addplot[color=red,]
    coordinates {
    (0, 17.5)
    (1, 11.7)
    (2, 9.2)
    (3, 7.8)
    (4, 7.2)
    (5, 6.5)
    (6, 6.3)
    (7, 6.1)
    (8, 5.9)
    (9, 5.8)
    (10, 6.0)
    };

\end{axis}
\end{tikzpicture}
\caption{\it WERs[\%] using soft distillation from non-causal teacher to a causal student. 
The alignment of the teacher model is shifted by N frames to the right.
The horizantal dashed line corresponds to the WER of FS distill without any shifting.
The WERs are computed by averaging over all dev and test sets of LibriSpeech.}
\label{fig:soft-shift}
\end{figure}
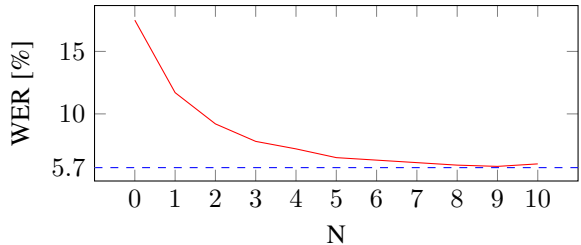

Experiments for comparing different KD methods for $\mathrm{LS}$ task using
non-causal and causal student models are shown in
\Cref{tab:libri-distill}.
In both cases, we can observe that FS distillation outperforms other distillation methods.
For non-causal student experiments, FS distillation achieves $27\%$ and $28\%$ relative
improvement in terms of WER on dev-other and test-other sets respectively 
compared to the student baseline model.
Moreover, FS distillation achieves $11\%$ and $8.3\%$ relative improvement compared to
hard distillation on dev-other and test-other sets respectively.
When using a causal student model, the RNN-T alignments of the student model
do not match the ones of the teacher model due to the lack of future context that
will delay the emission of output labels.
Thus, we can observe that soft distillation only works well when we shift the alignments
of the teacher to the right by N frames.
\Cref{fig:soft-shift} shows that we need to shift by 9 frames to achieve good
performance.
It also requires finding the correct frames shift.
FS distillation is robust towards mismatch of alignments and can already achieve better
results without any shifting.
Overall, we achieve $20\%$ and $18.9\%$ relative improvement in terms of WER on
dev-other and test-other sets respectively compared to the causal student model.

\begin{table}[t]
    \centering
    \caption{\it Comparing distillation methods from non-causal w2v-BERT XL teacher model to causal/non-causal student model on LibriSpeech dataset.
    WERs [\%] are reported on dev-other and test-other sets.}
    \begin{tabular}{|c|c|c||c|c|}
        \hline
         \multirow{2}{*}{Model} & \multicolumn{2}{c||}{Non-causal} & \multicolumn{2}{c|}{Causal} \\ \cline{2-5}
            & dev & test & dev & test \\ \hline
         Student & 4.4 & 4.6 & 10.4 & 9.5 \\ \hline \hline
         Hard & 3.6 & 3.6 & \phantom{0}9.4 & 8.6  \\ \hline
         Soft & 3.7 & 3.8 & \phantom{0}8.4 & 7.9 \\ \hline
         Hard + Soft & 3.7 & 3.8 & \phantom{0}8.6 & 8.2 \\ \hline
         Full-sum & \textbf{3.2} & \textbf{3.3} & \phantom{0}\textbf{8.3} & \textbf{7.7} \\ \hline
    \end{tabular}
    \label{tab:libri-distill}
\end{table}


\section{Large Scale Distillation}
\label{sec:production}
In this section, we demonstrate the robustness and scalabilty of full-sum distillation to models trained with several thousands of hours of labeled and unlabeled data in two Indic languages, Bengali and Malayalam. The in-house ASR training data comprises of short voice search utterances that are anonymized and hand-transcribed, and representative of Google’s voice search traffic. The supervised training data for Bengali and Malayalam contains 7.5M and 2.6M transcribed utterances which approximately corresponds to 9.4K and 4.7K hours respectively. This data is further augmented with various noise styles~\cite{kim2017generation}, time and frequency masking-based augmentation~\cite{specaug} and simulated multi-microphone utterances~\cite{Kim2017GenerationOL}.
The unsupervised training data consists of 59.8M utterances for Bengali and 23.5M utterances for Malayalam which approximately corresponds to 75K and 42.5K hours respectively. The development set is a small fraction of the training set held out for validation. 
The test set comprises of anonymous, transcribed utterances from the voice-search task (3.7k utterances for Bengali, 9.2k utterances for Malayalam).
We report error rates using the transliteration-optimized WER metric described in~\cite{emond2018transliteration} to accommodate mixed writing scripts frequently seen in Indics.

Both non-causal and causal models are chosen to act as teacher, while the student is kept as causal model. All the student models used in this experiment are initialized with the causal teacher model.
\Cref{tab:bnml-distill} shows that the full-sum (L1 with norm) provides consistent gains across both Bengali and Malayalam for both causal and non-causal teacher models.
We observed that the production data contains several outliers, as corroborated by a high WER of 16.6\% obtained using the full-sum MSE loss on Bengali with non-causal teacher.

\begin{table}[t]
    \centering
    \caption{\it Comparing distillation methods from 120M teacher (causal and non-causal) model to 120M causal student model.}
    \begin{tabular}{|c|c|c|c|c|}
        \hline
         \multirow{2}{*}{Model} & \multicolumn{2}{c|}{Bengali} & \multicolumn{2}{c|}{Malayalam} \\ \cline{2-5}
            & Causal & Non-causal & Causal & Non-causal \\ \hline
         Teacher & 16.4 & 13.3 & \textbf{33.4} & 32.5 \\ \hline \hline
         Hard & 15.4 & 15.1 & 37.5 & 35.2 \\ \hline
         Soft & 15.6 & 15.4 & 35.4  & 33.6  \\ \hline
         Full-sum & \textbf{15.0} & \textbf{14.9} & 33.9 & \textbf{33.3} \\ \hline
    \end{tabular}
    \label{tab:bnml-distill}
\end{table}

\section{Conclusions}

We investigated using sequence-level knowledge distillation (KD) methods, 
namely full-sum (FS) distillation, for recurrent neural network transducer (RNN-T) models for the first time.
We showed how to effectively distill knowledge from bad teacher models that
can generate noisy pseudo labels for training student models.
We also showed that FS distillation is robust towards discrepancies of RNN-T
alignments between teacher and student models.
We applied FS distillation on public data and large scale in-house production data, where it outperformed other KD methods.

\section{Acknowledgement}
We thank Dongseong Hwang, Gary Wang, Zhong Meng, Zhehuai Chen, Isabel Leal,  Oscar Chang for useful discussions.

\vfill\pagebreak



\renewcommand{\baselinestretch}{1.0}\normalsize
\let\OLDthebibliography\thebibliography
\renewcommand\thebibliography[1]{
  \OLDthebibliography{#1}
\let\normalsize\small\normalsize
  \setlength{\parskip}{4pt}
  \setlength{\itemsep}{2pt}
}

\bibliographystyle{IEEEbib}
\bibliography{strings,refs}

\end{document}